\documentclass{article}
\usepackage{ijcai}

\usepackage{times}
\usepackage{soul}
\usepackage{url}
\usepackage[hidelinks]{hyperref}
\usepackage[utf8]{inputenc}
\usepackage[small]{caption}
\usepackage{graphicx}
\usepackage{amsmath}
\usepackage{amsthm}
\usepackage{booktabs}
\usepackage{algorithm}
\usepackage{algorithmic}
\usepackage[switch]{lineno}

\usepackage{multirow}
\usepackage{indentfirst}

\usepackage{listings} 
\title{ALFRED: Ask a Large-language model For Reliable ECG Diagnosis}

\author{
    Jin Yu \and
    JaeHo Park\and
    TaeJun Park\and
    Gyurin Kim\and
    JiHyun Lee\and
    Min Sung Lee\and \\
    Joon-myoung Kwon\and
    Jeong Min Son\And
    Yong-Yeon Jo
\affiliations MedicalAI Co., Ltd.,  Seoul, Korea
\emails
\{jin.yu, wogh2012, skdf1020, gyu\_ll, rhdwn082, lylm, cto, jmson, yy.jo\}@medicalai.com
}

\begin{document}

\maketitle

\begin{abstract}
Leveraging Large Language Models (LLMs) with Retrieval-Augmented Generation (RAG) for analyzing medical data, particularly Electrocardiogram (ECG), offers high accuracy and convenience.
However, generating reliable, evidence-based results in specialized fields like healthcare remains a challenge, as RAG alone may not suffice.
We propose a Zero-shot ECG diagnosis framework based on RAG for ECG analysis that incorporates expert-curated knowledge to enhance diagnostic accuracy and explainability. 
Evaluation on the PTB-XL dataset demonstrates the framework's effectiveness, highlighting the value of structured domain expertise in automated ECG interpretation. 
Our framework is designed to support comprehensive ECG analysis, addressing diverse diagnostic needs with potential applications beyond the tested dataset.
\end{abstract}

\section{Introduction}

Recent advancements in Large Language Models (LLMs) have greatly improved the analysis of medical data, including Electrocardiogram (ECG), leading to automated and precise diagnostic tools. However, challenges remain, such as the persistent issue of hallucination in LLMs and the difficulty in generating clinically meaningful explanations. Retrieval-Augmented Generation (RAG) solves these by augmenting the prompt with content from medical books and related webpages, allowing the LLM to provide fact-based and reliable interpretations.

Despite its promise, RAG faces limitations if not grounded in verified expert knowledge, leading to potential diagnostic errors and oversimplified conclusions. This emphasizes the importance of integrating expert knowledge to ensure accurate and interpretable results in automated diagnostics.

We present a zero-shot diagnosis framework based on RAG, which combines expert-curated ECG documents and knowledge, a feature extraction model, and a rule module. Our evaluation using the PTB-XL dataset demonstrates that expert knowledge significantly enhances performance and provides detailed, accurate explanations. We have publicly released this framework as an application named "ALFRED: Ask a Large-language model For Reliable Electrocardiogram Diagnosis"~\footnote{\href{https://huggingface.co/spaces/MedicalAI-DP/Alfred}{https://huggingface.co/spaces/MedicalAI-DP/Alfred}}.


\begin{figure*}[tb!]
    \centering
    \includegraphics[width=0.85\linewidth]{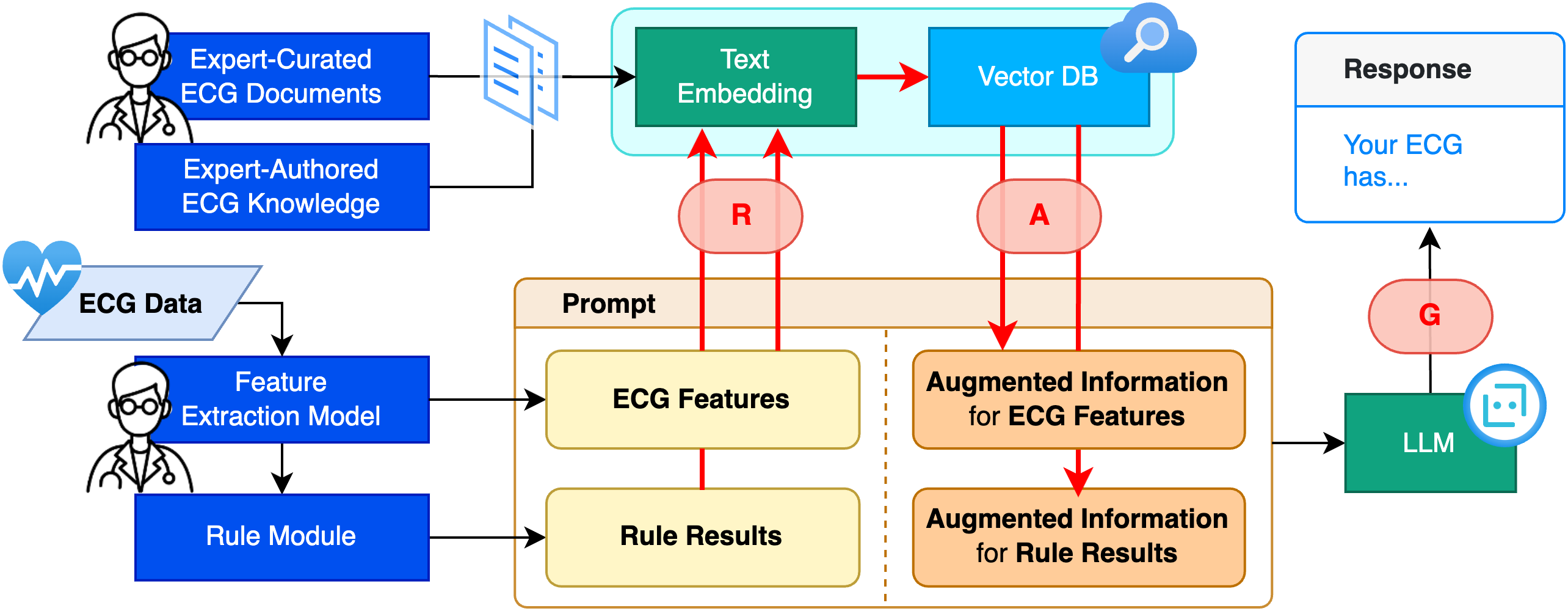}
    \caption{The proposed framework and its process pipeline.}
    \label{fig:framework}
\end{figure*}

\section{Framework}

Figure~\ref{fig:framework} illustrates our ECG analysis framework, which is based on RAG approach to enhance the prompt with expert knowledge retrieved from database \cite{yu2023zero}. The process begins by passing raw ECG input through the \textit{feature extraction module} to extract both lead-specific and global features. These features are then processed by the \textit{rule module} to generate diagnostic results using algorithm. Next, the extracted features and rule results are used to create a retrieval-augmented prompt, which is enriched with relevant information from a \textit{curated database}. This enriched prompt is provided to an LLM, generating a final diagnostic output guided by the extracted features, rule results, and curated reference material, offering medically meaningful and logical guidance.

\subsection{Components}

Our framework incorporates the following three key components to refine RAG results.
\paragraph{Curated Database:} 
Establishing a high-quality vector database is essential for effective RAG. 
Rather than indiscriminately collecting data, it's crucial to focus on carefully curated sources, vectorize them, and thoughtfully organize accumulated knowledge, especially in fields like medicine.
In our case, we construct (1) a core set of documents drawn from a broad range of ECG-related papers and reputable websites, with guidance from medical experts to ensure the inclusion of the most relevant material.
Additionally, beyond openly accessible documents, we incorporate (2) specialized knowledge by defining terms that clinicians consider critical for interpreting ECG.

\paragraph{Feature Extraction Module:}
Because clinicians typically begin by analyzing the positions of key waveform components—such as the P wave, QRS complex, and T wave—we designed our approach to align with this conventional workflow. 
To accomplish this, we develop a module that extracts features traditionally obtained from standard ECG devices (e.g., GE or Philips). 
We first implement a deep neural network-based segmentation model to delineate the ECG. 
Using the segmented results, we then perform further calculations to derive both lead-specific and global features.

\paragraph{Rule Module:}
In the subsequent stage of the workflow, clinicians combine the extracted features according to disease-specific criteria to reach a diagnosis. 
Therefore, we develop a rule module that incorporates this diagnostic logic (i.e., Rule). 
This module is capable of analyzing any condition that can be diagnosed using ECG data.




\subsection{Prompt Engineering}

Our prompt is composed of the following five main parts: 
\textsf{(1) ECG Features:} It includes both lead-specific and global features. The number of lead-specific features, such as \textit{PR interval and amplitude of R waves}, is 30, and the number of global features, such as the \textit{heart rate and corrected QT interval}, is 12, providing a comprehensive representation of the ECG. Additionally, subject-specific information such as \textit{age} and \textit{gender} is also included, which are essential for accurate ECG interpretation.
\textsf{(2) Rule Results:} It generates a list of diagnosed conditions based on ECG features, covering 40 categories, with each disease marked as True/False for presence or absence. This includes warnings related to ECG quality as well as a variety of cardiac conditions such as bundle branch block and atrial fibrillation. 
\textsf{(3) Augmented Information for ECG Features:} This part involves retrieval-augmented content based on ECG feature names as queries. For example, for the \textit{PR interval}, content retrieved by a query such as \textit{"PR interval is defined as something or other"} is added to the prompt.
\textsf{(4) Augmented Information for Rule Results:} This part refers to retrieval-augmented content generated by querying the disease names included in the rule results. For example, for the \textit{MI}, content retrieved by a query such as \textit{"Myocardial Infarction is defined as something or other"} is added to the prompt.
\textsf{(5) Instruction \& Response Format:} This part includes instructions on how to read ECG and a directive for the LLM to respond in a specific format. Additionally, it includes a request to draw conclusions based on the four main parts above.


\section{Evaluation}

\begin{table*}[tb!]
\centering
\begin{tabular}{c|c|cc|c|c|c|c}
    \toprule
        \multirow{2}{*}{\textbf{Framework}} & \multirow{2}{*}{\textbf{Rule Results}} & \multicolumn{2}{c|}{\textbf{Augmented Information}} & \multirow{2}{*}{\textbf{PPV}} & \multirow{2}{*}{\textbf{NPV}} & \multirow{2}{*}{\textbf{Sens.}} & \multirow{2}{*}{\textbf{Spec.}} \\
         &  & \multicolumn{1}{c|}{\textbf{ECG Features}} & \textbf{Rule Results} &  &  &  &  \\
    \midrule
        \textbf{Base} & $\times$ & \multicolumn{1}{c|}{Documents only} & Documents only$\mathbf{^*}$ & 0.326 & 0.805 & 0.356 & 0.754 \\
        \textbf{Ablation1} & $\bigcirc$ & \multicolumn{1}{c|}{Documents only} & Documents only & 0.414 & 0.820 & 0.458 & 0.761 \\
        \textbf{Ablation2} & $\bigcirc$ & \multicolumn{1}{c|}{Documents + Knowledge} & Documents only & 0.416 & 0.819 & 0.454 & 0.763 \\
        \textbf{Ablation3} & $\bigcirc$ & \multicolumn{1}{c|}{Documents only} & Documents + Knowledge & \textbf{0.451} & \textbf{0.834} & \textbf{0.477} & \textbf{0.798} \\
        \textbf{Proposed} & $\bigcirc$ & \multicolumn{1}{c|}{Documents + Knowledge} & Documents + Knowledge & 0.443 & \textbf{0.834} & \textbf{0.477} & 0.797 \\
    \bottomrule
\end{tabular}
\begin{minipage}{0.95\textwidth}
\footnotesize{* In the Base, there are no rule results, but since the disease to be diagnosed is predefined in the PTB-XL dataset, retrieval is possible.}
\end{minipage}
\caption{Performance comparison on framework configurations}
\label{tab:experiment_design}
\end{table*}

\subsection{Experimental Settings}

\paragraph{Dataset}
To evaluate our proposed framework, we utilized the PTB-XL(v1.0.3) dataset\cite{ptbxl2020}. The PTB-XL dataset contains 21,799 12-lead ECG recordings collected from 18,869 patients. 
To ensure consistency and fairness in evaluation, the dataset is stratified into 10 folds. For our experiments, we only used the 10th fold with a particularly high label quality.
Each ECG recording in the dataset was annotated by up to two cardiologists. 
Annotations are categorized into three types: diagnostic, form, and rhythm statements. We used only the diagnostic statements as labels for evaluation. 
Each diagnostic statement has a likelihood ranging from 0 to 100, extracted from ECG reports based on relevant keywords. 
Diagnostic statements are further aggregated into five superclasses: Normal  (NORM), Conduction Disturbance (CD), Hypertrophy (HYP), Myocardial Infarction(MI), and ST-T Changes (STTC). 
For each superclass, it is labeled as 1 if at least one corresponding diagnostic statement has a likelihood score of 50 or higher; otherwise, it is labeled as 0.

\paragraph{Implementation}
Various techniques and deep learning models form the foundation of our framework.
We leverage Milvus~\footnote{https://milvus.io} as the underlying vector database, where the input textual data is split into chunks of up to 1,024 characters while ensuring sentence integrity. Additionally, we use the following index parameters for efficient indexing and retrieval: (1) similarity metric used to measure similarities among vectors: COSINE (2) indexing algorithm: HNSW with parameters (efConstruction: 100, M: 10).
Each chunk is then transformed into a high-dimensional embedding space using OpenAI's text-embedding-3-large model~\footnote{https://platform.openai.com/docs/guides/embeddings}, enabling efficient construction and retrieval of relevant information. 
We structured the database used for information retrieval into two distinct types: (1) a collection of textbooks and web pages, referred to as \textit{documents}, and (2) expert-authored professional contents, referred to as \textit{knowledge}.

For feature extraction, we adopt a UNet~\cite{ronneberger2015unet} and carefully tuning hyperparameters such as network depth and the number of convolutional layers at each level to optimize the model for ECG delineation tasks. ECG features are processed through our feature-based algorithm to list potential diseases. To ensure consistency with the experimental settings, we mapped disease name from rule results to align with the five superclasses of the PTB-XL dataset.
ECG features and diagnosis names are used to search for the definitions of each term in the previously constructed vector database, creating the final prompt.
These tailored prompts are subsequently processed by OpenAI's GPT-4o-Mini model\footnote{version: 2024-07-18}, which generates coherent and contextually accurate final responses.

\paragraph{Comparisons}
To evaluate the impact of our framework, we conducted a comparative evaluation using multiple configurations. 
The feature extraction module is included in every configuration.
There are three configurable options: (1) the inclusion of rule results, (2) the use of augmented information for features, and (3) the use of augmented information for rule results—then consider formatting them more clearly.
The augmented information can be derived either solely from documents or from both documents and knowledge.
By selectively activating these options, we systematically examined their individual and combined effects on performance.



\subsection{Experimental Results}

\paragraph{Classification performance:}
Table~\ref{tab:experiment_design} presents a performance comparison across various configurations, ranging from the Base to the Proposed framework, by systematically ablating options. Performance is evaluated across four metrics, with the best results highlighted in bold for emphasis.

The results indicate a gradual increase across all metrics from the Base to the Proposed framework. 
Due to the inclusion of \textit{Rule Results}, Ablation1 was a significant improvement in all metrics from Base. It appears that the rule results served as hints for the LLM in ECG interpretation, leading to the enhanced performance.
The next most notable performance improvement occurred with the addition of our \textit{Knowledge} to the Augmented Information of Diagnosis. It seems that the complementary information from both documents and knowledge refined the basis for ECG interpretation, leading to improved performance.
Compared to Ablation3, PPV and Specificity (Spec) slightly decreased in the proposed configuration, likely due to the increased amount of prompts, which may have caused confusion for the LLM in making judgments.

By comparing both Base and Proposed frameworks, we demonstrate that our clinical expertise have contributed to performance improvements within this proposed framework for diagnosing ECG using LLM. 

\begin{lstlisting}[label={exam:exam}, captionpos=b, caption={An excerpt from the LLM output for samples with some diseases (ecg\_id: 21,764 in PTB-XL(v1.0.3)).}]
{
  "CD": {
    "Result": true,
    "Explanation": "The QRS duration is prolonged (median of 113 ms), indicating a conduction disturbance. The presence of a QS pattern in leads V1 and V2 further supports this diagnosis."
  },  // Other classes are omitted.
}
\end{lstlisting}


\paragraph{Interpretability:}
In addition to classifying each superclass, we provided explanation examples to qualitatively evaluate the responses generated by the LLM, as shown in Example~\ref{exam:exam}.
Our in-house medical experts found the explanations to be highly effective and well-constructed. 
They particularly appreciated the LLM's ability to identify relevant features and present diagnostic conditions in a clear and straightforward manner. 
This approach not only highlighted key factors for each diagnosis but also used accessible language, making the explanations easy to understand, even for non-experts.

\section{Conclusion and Limitations}

We developed a RAG-based zero-shot Diagnosis framework to analyze ECGs, integrating expert-curated knowledge to address the complexities of medical diagnostics.
Through close collaboration with clinicians, we established a well-curated vector database and developed a feature extraction module as well as a rule module incorporating feature-based diagnostic methods.
This approach ensures results that are both reliable and aligned with clinical standards.
Future work will focus on enhancing LLM integration, including optimized embedding models, advanced tuning techniques, and real-world dataset evaluation to address greater variability. 
By combining technical advancements with clinical expertise, we aim to further refine the framework for broader applicability in real-world medical settings.

\bibliographystyle{named}
\bibliography{ijcai}

\end{document}